\documentclass[10pt,twocolumn,letterpaper]{article}

\usepackage{cvpr}
\usepackage{times}
\usepackage{epsfig}
\usepackage{graphicx}
\usepackage{amsmath}
\usepackage{amssymb}

\usepackage[breaklinks=true,bookmarks=false]{hyperref}

\cvprfinalcopy 

\def\cvprPaperID{***} 

\begin{document}

\title{A Feasible Framework for Arbitrary-Shaped Scene Text Recognition \\
\large Technical Report}


\author{Jinjin Zhang$^{1}$, Wei Wang$^{2}$, Di Huang$^{1}$, Qingjie Liu$^{1}$ and Yunhong Wang$^{1}$\\
$^{1}$Beihang University\\
$^{2}$Institute of Automation, Chinese Academy of Sciences (CASIA)\\
{\tt\small zhang0jhon@gmail.com, wangwei@nlpr.ia.ac.cn, \{dhuang, qingjie.liu, yhwang\}@buaa.edu.cn}
}

\maketitle

\begin{abstract}
   Deep learning based methods have achieved surprising progress in Scene Text Recognition (STR), one of classic problems in computer vision. In this paper, we propose a feasible framework for multi-lingual arbitrary-shaped STR, including instance segmentation based text detection and language model based attention mechanism for text recognition. Our STR algorithm not only recognizes Latin and Non-Latin characters, but also supports arbitrary-shaped text recognition. Our method wins the championship on Scene Text Spotting Task (Latin Only, Latin and Chinese) of ICDAR2019 Robust Reading Challenge on Arbitrary-Shaped Text Competition. Code is available at \url{https://github.com/zhang0jhon/AttentionOCR}.
\end{abstract}

\section{Introduction}

Deep learning has brought significant revolution in computer vision and machine learning in recent years. Since AlexNet~\cite{krizhevsky2012imagenet} won the championship on ImageNet~\cite{deng2009imagenet} image classification in 2012, an increasing number of researchers have paid attention to deep learning and its wide application, such as computer vision, natural language process and speech recognition. With the rapid development of deep learning, tremendous improvement has been achieved in various research areas, especially in machine learning and artificial intelligence. For instance, Batch Normalization~\cite{ioffe2015batch} and ResNet~\cite{he2016deep} make it possible to train deeper neural networks steadily as well as alleviate gradient vanishing and exploding problems. R-CNN based object detection series~\cite{girshick2014rich, girshick2015fast, ren2015faster} significantly improve mean Average Precision (mAP) and Mask R-CNN~\cite{he2017mask} extend R-CNN based detection methods to instance segmentation. Semi-supervised attention mechanism in deep learning demonstrates its effectiveness in both computer vision~\cite{mnih2014recurrent, xu2015show, hu2018squeeze} and natural language process~\cite{bahdanau2014neural, luong2015effective, vaswani2017attention}. Moreover, Neural Architecture Search (NAS) ~\cite{zoph2016neural, liu2018progressive} can automatically find the optimal network in various areas, including EfficientNet~\cite{tan2019efficientnet}, EfficientDet~\cite{tan2019efficientdet}, and so on.

\begin{figure}[t]
\begin{center}
\includegraphics[width=0.9\linewidth]{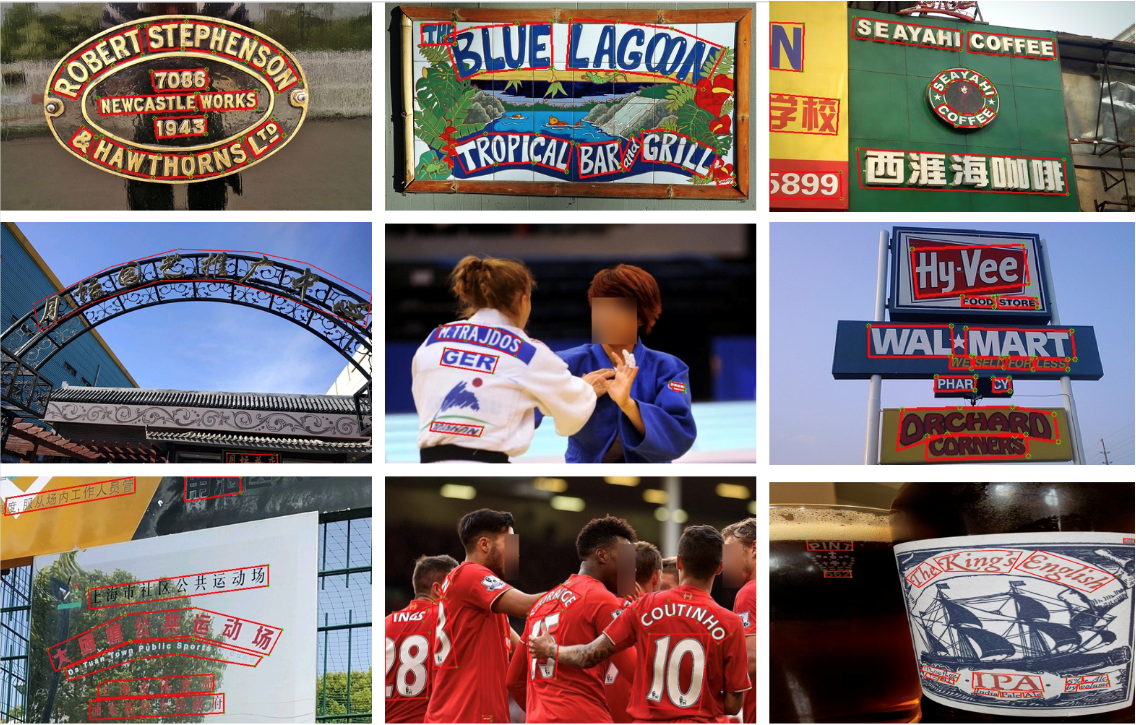}
\end{center}
   \caption{Challenges in STR. Images from ICDAR2019-ArT~\cite{chng2019icdar2019} show the STR challenges in real-world scenes including various languages and shapes.}
\label{fig:challenges}
\end{figure}

STR, one of the extensively used techniques, benefits a lot from the boom of deep learning. Nevertheless, there are still several problems in realistic scenario as show in Fig. \ref{fig:challenges}. Firstly, various text regions including horizontal, vertical and curved text require a high-quality detector to handle with arbitrary-shaped text detection. Secondly, it is difficult for a general model to recognize text region with a variety of languages and shapes.

\begin{figure*}
\begin{center}
\includegraphics[width=0.9\linewidth]{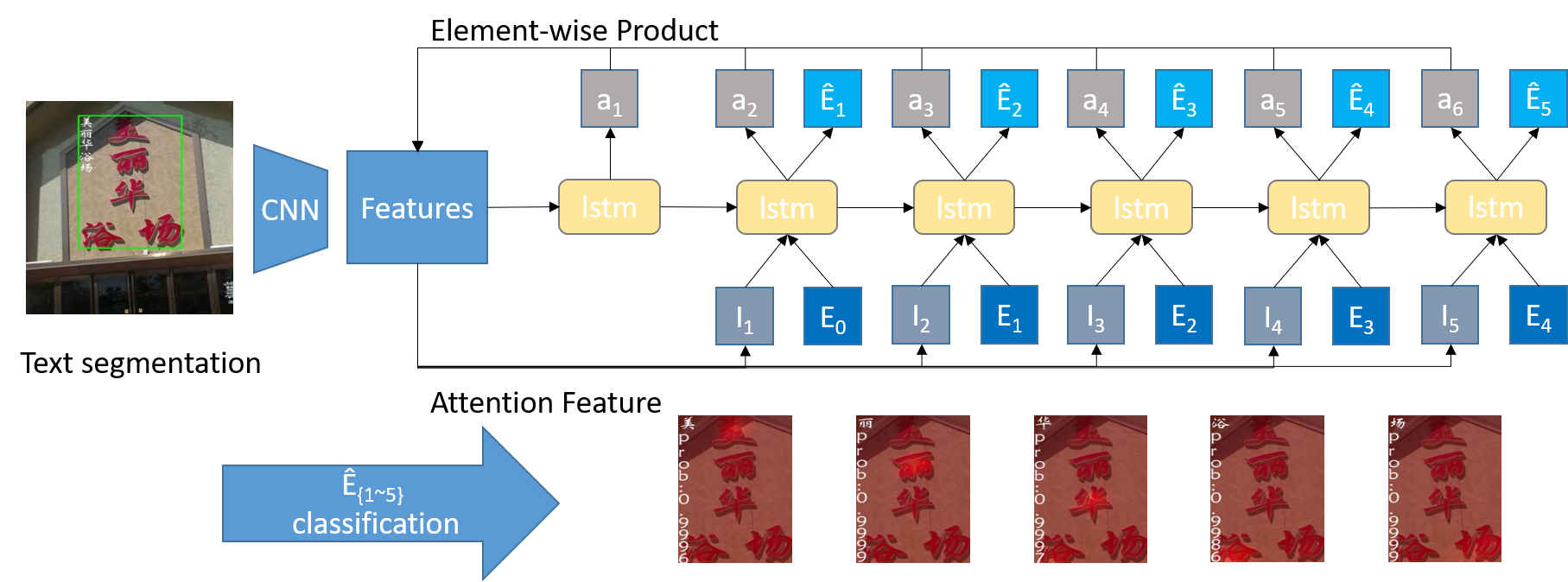}
\end{center}
   \caption{STR architecture. Our algorithm first conduct text segmentation with Cascade Mask R-CNN and then recognize text region with attention based sequence transduction method.}
\label{fig:architecture}
\end{figure*}

In this paper, we deem STR as a cross domain problem including both object detection and language model based sequence transduction instead of simple text detection and classification, which benefits from the recent achievements in computer vision and natural language process. Instead of individually concentrating on scene text detection or text recognition, we propose a feasible framework for multi-lingual arbitrary-shaped scene text spotting~\cite{chng2019icdar2019}. Our algorithm adopts general instance segmentation method for robust arbitrary-shaped text detection, and simultaneously takes context information into consideration for text recognition by attention based Word2vec~\cite{mikolov2013efficient} method. Furthermore, word embedding based arbitrary-shaped text recognition guarantees the convergence of end-to-end soft attention mechanism via a weakly supervised method. In brief, we present an universal and robust proposal for real-world STR by the combination of instance segmentation and attention based sequence transduction methods.

The main contributions in this paper consist of three aspects as following:

1. We propose a feasible framework for STR that is capable of multi-lingual arbitrary-shaped text detection and recognition, and the results in several STR datasets demonstrate the effectiveness and robustness of our algorithm.

2. We propose an attention based text recognition method which is able to recognize irregular text in Latin or Non-Latin characters in the unique model and combine with the language model based method.

3. Our algorithm is easy to extend with state-of-the-art algorithm in attention mechanism for high level applications such as text based visual question answering, semantic parsing, et al.


\section{Related Work}

In this section, we will present the recent work relevant to STR, including object detection, attention mechanism for alignment in sequence transduction. Our STR framework is designed referring to previous work introduced below, aiming to handle STR in a general way which combines object detection and attention mechanism techniques.

\subsection{Text Detection}

Object detection is one of the fundamental research areas in computer vision which has attracted extensive attention with the remarkable progress in the last several years.  There have been many efficient object detector architectures, such as one-stage detector like YOLO~\cite{redmon2016you, redmon2017yolo9000, redmon2018yolov3}, SSD~\cite{liu2016ssd}, RetinaNet~\cite{lin2017focal}, as well as anchor-free detectors~\cite{law2018cornernet, duan2019centernet, tian2019fcos}. Nevertheless, the majority of one-stage detectors suffer from worser behavior on small objects and lower precision compared with two-stage object detectors. Two-stage detectors have better performance on realistic scenes in spite of higher computational requirement. Further improved architectures have been proposed to achieve robust and accurate results, like FPN~\cite{lin2017feature, ghiasi2019fpn}, DCN~\cite{dai2017deformable, zhu2019deformable}, SNIP~\cite{singh2018analysis, singh2018sniper}, Cascade R-CNN~\cite{cai2018cascade}, etc.

Regarding text detection in STR, multi-scale and irregular text regions are the principal problems appearing in real-world scenes. EAST~\cite{zhou2017east} and FOTS~\cite{liu2018fots} proposed for text detection still have limitation in arbitrary-shaped text.  Mask TextSpotter~\cite{lyu2018mask} adopts Mask R-CNN based instance segmentation method for end-to-end trainable scene text spotting but need character level segmentation. In this paper, we utilize Cascade R-CNN based irregular text region segmentation for accurate text localization which assists the following text recognition section.

\subsection{Text Recognition}

Connectionist Temporal Classification (CTC)~\cite{graves2006connectionist} is a widely used method in text recognition without knowing the alignment between the input and the output. Covolutional Recurrent Neural Network (CRNN)~\cite{shi2016end}, one of the most well-known methods in text recognition, is a combination of Convolution Neural Network (CNN), Recurrent Neural Network (RNN) and CTC loss for image-based sequence recognition tasks. However,  CTC based method is incapable of handling multi-oriented text because of single-oriented feature slice. With the advent of attention mechanism in deep learning, ~\cite{xu2015show} demonstrates the effectiveness of the attention based method for alignment between visual feature and word embedding in image caption.

Attention mechanism has been brought in deep learning since 2014, including reinforcement learning based Recurrent Attention Model (RAM)~\cite{mnih2014recurrent} in computer vision and end-to-end trainable Bahdanau Attention (Additive Attention)~\cite{bahdanau2014neural} which solves the context alignment problem existing in Seq2Seq~\cite{sutskever2014sequence} respectively. Luong Attention (Multiplicative Attention)~\cite{luong2015effective} concludes and extends Bahdanau Attention to general attention mechanism formulated by 3 steps: score function for similarity metric, alignment function for attention weight, and context function for aligned context feature. ~\cite{vaswani2017attention} proposes the Transformer architecture based solely on multi-head attention mechanism motivated by self-attention representation learning~\cite{cheng2016long} as well as end-to-end memory network~\cite{sukhbaatar2015end}, and simultaneously implements the parallelization in Transformer as CNN based methods like ByteNet~\cite{kalchbrenner2016neural}, ConvS2S~\cite{gehring2017convolutional}.

Inspired by the practicable end-to-end soft attention mechanism, we propose an attention based text recognition method for arbitrary-shaped text region segmented by Cascade Mask R-CNN, which aligns visual context with corresponding character embedding in an semi-supervised optimization method.

\section{Architecture}

As illustrated in Fig. \ref{fig:architecture}, with Cascade Mask R-CNN~\cite{cai2018cascade, he2017mask} based instance segmentation method, we extract the text region from image at first and then feed the masked image to attention based text recognition module for sequence classification.

\subsection{Cascade Mask R-CNN with Text-Aware RPN}

We adopt Cascade R-CNN~\cite{cai2018cascade} based instance segmentation method for text detection, but there is still slightly different from that in object detection. Because the majority of text regions are long narrow rectangles, it is inappropriate that applies conventional anchor-based two-stage object detector to text detection straight forward.

\begin{figure}[t]
\begin{center}
\includegraphics[width=0.9\linewidth]{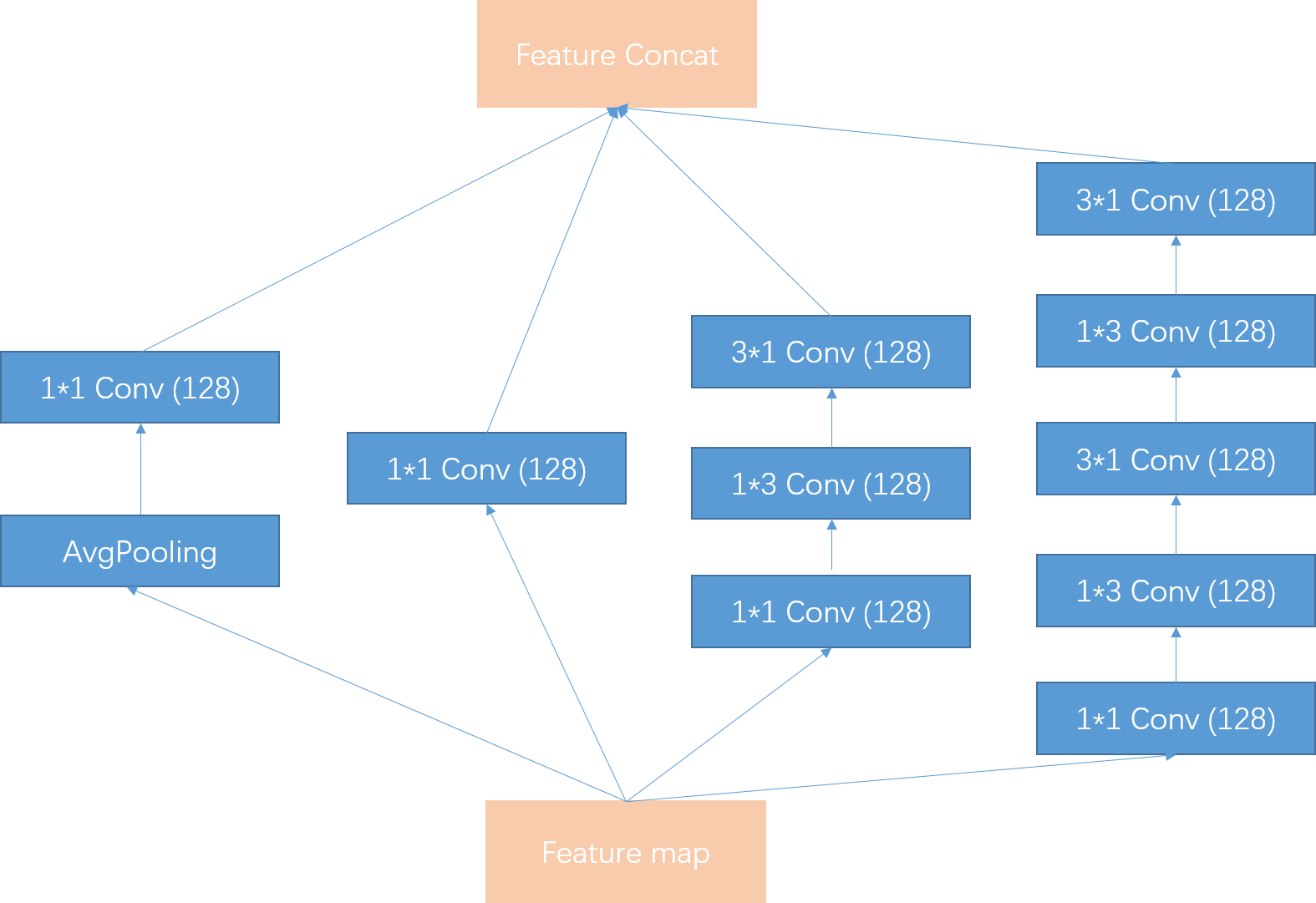}
\end{center}
   \caption{Inception text-aware RPN architecture. The architecture is adopted for text detection inspired from inception structure.}
\label{fig:inceptionrpn}
\end{figure}

With Faster R-CNN~\cite{ren2015faster} default positive-anchor parameter with Intersection over Union (IOU) threshold higher than 0.7, there are few positive anchors for text due to unmatched shapes and may induce difficult convergence in Region Proposal Network (RPN) during training. It is essential to redesign prior anchors sample strategy with proper anchor ratios and scales for more matched ground-truth positive anchors, namely True Positives (TP). Furthermore, inspired by recent architectures which demonstrate the effectiveness of $1 \times n$ and $n \times 1$ kernels including InceptionV3~\cite{szegedy2016rethinking}, TextBoxes~\cite{LiaoSBWL17, Liao2018Text}, an inception text-aware RPN architecture as shown in Fig. \ref{fig:inceptionrpn} is proposed for rich and robust text feature extraction. Cascade R-CNN, a multi-stage extension of two-stage R-CNN object detection framework, is exploited to obtain high quality object detection, which can effectively reject close False Positives (FP) and improve the detection performance. Finally, The mask branch as in Mask R-CNN~\cite{he2017mask} in the text detection module predicts the final masked arbitrary-shaped text regions for latter scene text recognition.

Regarding Cascade Mask R-CNN for text segmentation, we optimize the model by minimizing the multi-task loss function as below:
\begin{eqnarray}
L_{total} = L_{rpn} + L_{cascade} + L_{mask} + \lambda L_{reg} \\
L_{cascade} =  \sum_{i=1}^{N} ( L_{cls}(\hat{y}_{i}, {y}_{i}) + L_{loc}(\hat{b}_{i}, {b}_{i}) ) 
\end{eqnarray}
where N is the number of multiple cascade stages. $\hat{y}_{i}$ is label logits and ${y}_{i}$ is ground-truth one-hot labels. $\hat{b}_{i}$ is estimated bounding box transformation parameters and ${y}_{i}$ is the ground-truth. $L_{rpn}$ and $L_{mask}$ are the same as Mask R-CNN~\cite{he2017mask}. $L_{cascade}$ is the summation of multi-stage $L_{cls}$ and $L_{loc}$ with increasing IOU thresholds which represent the cross-entropy loss and smoothed $L_{1}$ loss in Fast R-CNN~\cite{girshick2015fast} respectively. $\lambda$ is the weight decay factor and $L_{reg}$ represents the $L_{2}$ regularization loss.

\subsection{Attention based Cross-Modal Alignment and Sequence Classification}

Regarding text recognition, it should be a cross-domain representation learning and alignment problem including computer vision and sequence transduction instead of simple image classification as far as we are concerned. The reason is that text recognition in realistic scenes always represents the linguistic significance. Namely, scene text recognition should benefit from language model based natural language process methods.

We aim to propose a universal sequence classification framework that is capable of integrating multi-lingual arbitrary-shaped text recognition in the unique recognition model. Long Short-term Memory (LSTM)~\cite{hochreiter1997long} born for sequence alignment is adopted for text sequence recognition in our framework. Furthermore, for purpose of irregular text recognition, we propose an visual attention based method referring to Bahdanau Attention Mechanism~\cite{bahdanau2014neural} which learns cross-modal alignment between visual feature and LSTM hidden state for classification in a semi-supervised way.

With the global word embedding matrix $E$ and CNN feature map $V$, we can model the sequence classification as soft end-to-end attention mechanism based alignment problems between visual attention based context feature $c_{t}$ from $V$ by attention weight $\alpha_{t}$ and the corresponding embedding $e_{t}$ in $E$ at each time step $t$. Let $C = c_{1}, c_{2}, ..., c_{t}$ be the context features and let $E = e_{0}, e_{1}, ..., e_{t}$ be the embeddings of $T$ time steps in the target string. Using the chain rule the conditional probability of the sequence $P(E | C)$ can be decomposed as as following:
\begin{eqnarray}
P(E | C) = \prod_{t=1}^{T} P(e_{t} |  h_{t-1}, c_{t}, \bf{\hat{e}_{t-1}}) 
\end{eqnarray}
where $T$ represents the maximum sequence length with the End of Sequence (EOS) character proposed in machine translation and $\bf{\hat{e}_{t-1}}$ is the \textbf{expectation} of previous embedding. Simultaneously $h_{t-1}$ is the previous LSTM hidden state and $c_{t}$ is the context vector calculated according to the following attention mechanism formulas:
\begin{eqnarray}
&\alpha_{t} = AttentionFunction(V, h_{t-1})   \\ 
&c_{t} = \sum \alpha_{t} \cdot V   ~~~~~~~~     1 \leq t \leq T \nonumber
\end{eqnarray}
where the initial LSTM hidden state $h_{0}$ in computed from $V$ representing global feature and the LSTM architecture is the same as that in image caption~\cite{xu2015show}. In addition, we choose Bahdanau Attention Mechanism~\cite{bahdanau2014neural} as our default $AttentionFunction$ formulated as following:
\begin{eqnarray}
e_{ij}^{t} = W_{e} \cdot tanh(W_{h}^{\alpha} \cdot h_{t-1} + W_{v} \cdot V) \\
\alpha_{t} = \frac{exp(e_{ij}^{t})}{\sum_{i=1}^{h} \sum_{j=1}^{w} exp(e_{ij}^{t})}
\end{eqnarray}
where $h$ and $w$ are the CNN feature map height and width respectively.

The final optimization function, namely Masked Cross Entropy (MCE) loss for dynamic sequence length, is formulated as below:
\begin{eqnarray}
MCE(Y|X;M) = \frac{-\sum_{t=1}^{T} m_{t} \cdot p_{t}log(q_{t})}{\sum_{i=1}^{T} m_{t}}
\end{eqnarray}
where $q_{t}$ is logits calculated from previous embedding $\bf{\hat e_{t-1}}$, current context feature $c_{t}$ and LSTM hidden state $h_{t}$ as below:
\begin{eqnarray}
q_{t} \propto W_{h}^{e} \cdot h_{t} + W_{c} \cdot c_{t} + W_{e} \cdot \bf{\hat{e}_{t-1}}
\end{eqnarray}
$p_{t}$ is the one-hot ground-truth sequence labels, $m_{t}$ is the corresponding sequence mask for dynamic sequence labels as following:
\begin{eqnarray}
m_{t} = \left\{
\begin{aligned}
1 & ~~~~ x_{t} ~ is ~ not ~ EOS ~ or ~ the ~ first ~ EOS, \\
0 & ~~~~ otherwise. \\
\end{aligned}
\right.
\end{eqnarray}


\section{Experiment}

\begin{figure}[t]
\begin{center}
\includegraphics[width=0.9\linewidth]{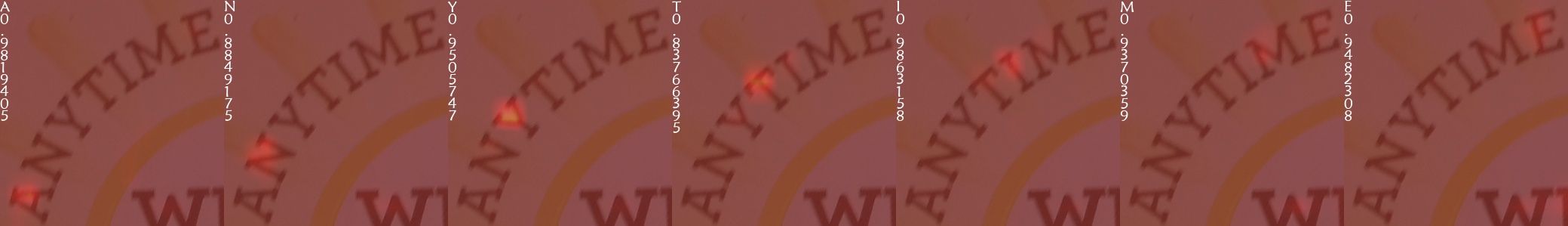}
\includegraphics[width=0.9\linewidth]{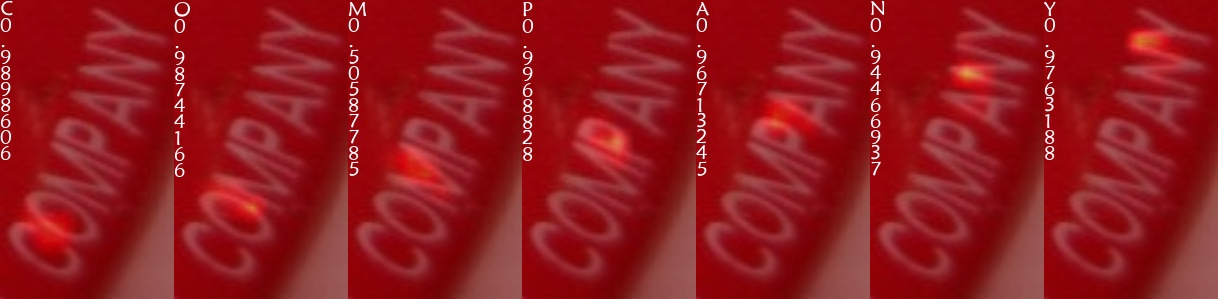}
\end{center}
   \caption{Attention mechanism visualization. The results show the correct attention location in sequence and demonstrate the effectiveness of attention mechanism based arbitrary-shaped text recognition.}
\label{fig:attention}
\end{figure}

In this section, we will present our instance segmentation based text detection and attention based text recognition and how to train the models in detail. We are not able to build an end-to-end STR framework due to the limited 1080Ti GPU memory. Our STR algorithm is implemented based on TensorFlow~\cite{abadi2016tensorflow} and its high-level API Tensorpack~\cite{wu2016tensorpack}, with speed and flexibility built together.

\subsection{Text Segmentation}

We follow the default protocol in Mask R-CNN and Cascade R-CNN with ResNet101-FPN backbone besides some modified parameters for text detection. We use RPN batch size 128 and batch size 256 for ROIAlign head, half as the default Mask R-CNN parameters due to the less TP anchors in text detection. Note that for the purpose of data augmentation, we conduct random affine transformation to generate more text images including rotate, translate, shear, resize and crop transform.

Stochastic Gradient Descent (SGD) with momentum is adopted as default text segmentation optimizer, with learning rate of 0.01 and momentum of 0.9. We use a weight decay of 0.0001 and train on 8 GPUs with 1 image per GPU for 360k iterations,  with the learning rate decreased by 10 at the 240k iteration and 320k iteration. We set the maximum image size to 1600 for accurate small text region. Training images are resized for multi-scale training with their scale (shorter edge) in the range of 640 to 1280 pixels. Pre-train detection model from ICDAR2017-MLT~\cite{nayef2017icdar2017}, ICDAR2017RCTW~\cite{shi2017icdar2017} is used for weight initialization for text detection. Furthermore, the RPN anchor ratios of [1/9, 1/4, 1/2, 1, 2, 4, 9] is adopt for extremely irregular text detection. Then the mask branch predicts the final segmented text region for latter text recognition.

\subsection{Sequence Classification}


\begin{figure}[t]
\begin{center}
\includegraphics[width=0.9\linewidth]{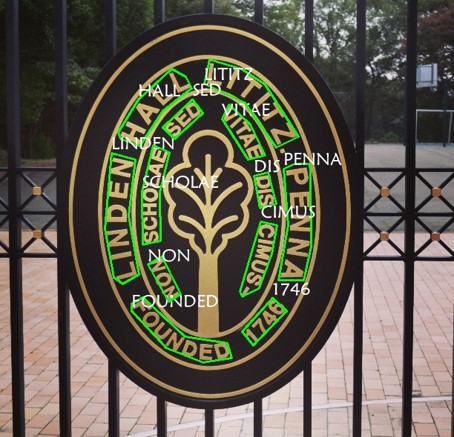}
\end{center}
   \caption{Arbitrary-oriented text detection and recognition.}
\label{fig:example}
\end{figure}

\begin{figure*}
\begin{center}
\includegraphics[width=0.9\linewidth]{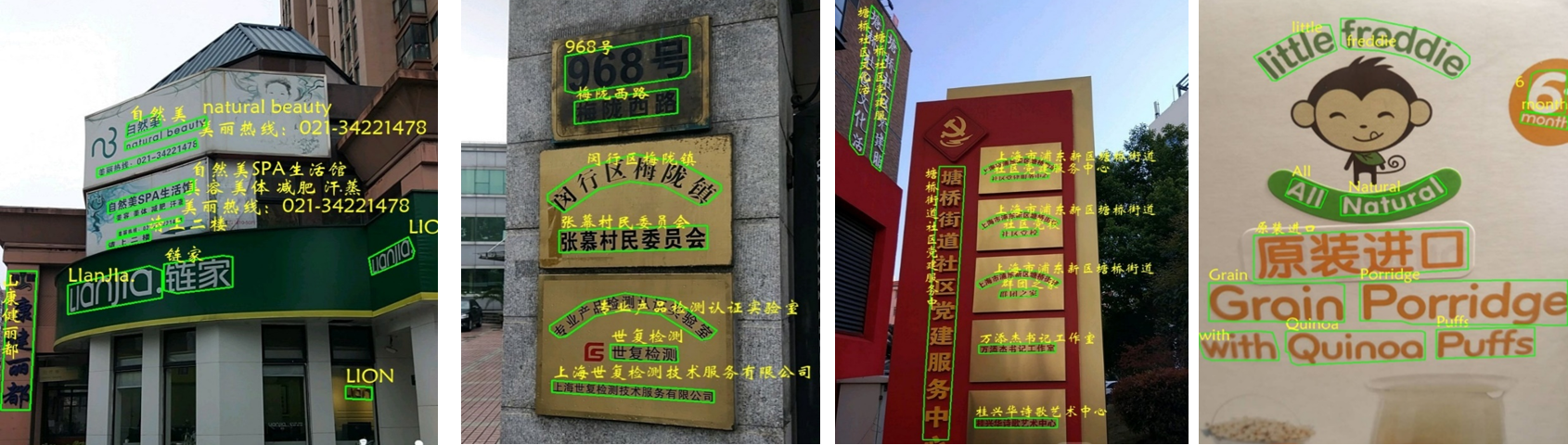}
\end{center}
   \caption{STR visualization. The results prove that our framework is instrumental for multi-lingual arbitrary-shaped STR and simultaneously robust for realistic scenes.}
\label{fig:viz}
\end{figure*}

In this part, we exploit InceptionV4~\cite{szegedy2017inception} as our text recognition backbone, which is appropriate for fine-grained distinguishable text feature extraction. Furthermore, referring to Google's Neural Machine Translation (GNMT)~\cite{wu2016google, luong17} and image caption methods\cite{xu2015show}, we model the text recognition as sequence transduction between visual feature and sequence embedding by Bahdanau Attention based LSTM. In addition, EOS and sequence mask are adopted for sequence padding to transform various length sequences to fixed length.

For the purpose of universal multi-lingual arbitrary-shaped text recognition, we conduct random affine transformation as well as random color jittering with text image and corresponding polygons and then crop and resize image with longer size of 256 pixels. Finally, we random translate and flip augmented image and pad the irregular text region to size of $299 \times 299$ with zero. The fixed maximum sequence length, namely LSTM sequence length, is set as 26 ($T$ of 27 indeed with EOS). Compared with conventional LSTM, our LSTM takes not only previous hidden state and cell state, but alse the current attention feature and previous embedding as input. The LSTM size is 512 and embedding matrix shape is $5435 \times 256$ representing embedding size of 256 and 5435 characters including Chinese, English letters, digits and other symbols.

We train our recognition model via Adam~\cite{kingma2014adam}, an adaptive learning rate optimizer, with learning rate of 0.0001. We use a weight decay of 0.00001 and train with 24 image batch size on 1 GPU for 460k iterations, with the learning rate decreased by 10 at the 320k iteration and 400k iteration. Pre-train InceptionV4 model from ImageNet is loaded for initialization and multiple datasets including LSVT, ArT, ReCTS, COCO-Text, ICDAR2017 are used to train the multi-lingual irregular text recognition model. Meanwhile, the padded fixed length sequences with corresponding sequence masks in GNMT make it possible to compute masked cross entropy loss for dynamic sequence length.

As shown in Fig. \ref{fig:attention}, our algorithm is capable of learning the alignment and classifying sequence in a simi-supervised method. Meanwhile, Fig. \ref{fig:example} demonstrate that our model is able to handle arbitrary-oriented text recognition.

\subsection{Results}

\begin{table}
\begin{center}
\begin{tabular}{|l|c|}
\hline
Recall & 49.29\% \\
Precision & 56.03\% \\
Hmean & 52.45\% \\
\hline\hline
1-N.E.D & 53.86\% \\
\hline
\end{tabular}
\end{center}
\caption{STR Results on ICDAR2019-ArT (Latin Only).}
\label{table:result_latin}
\end{table}

\begin{table}
\begin{center}
\begin{tabular}{|l|c|}
\hline
Recall & 47.98\% \\
Precision & 52.56\% \\
Hmean & 50.17\% \\
\hline\hline
1-N.E.D & 54.91\% \\
\hline
\end{tabular}
\end{center}
\caption{STR Results on ICDAR2019-ArT (Latin and Chinese).}
\label{table:result_latin_and_chinese}
\end{table}

\begin{table}
\begin{center}
\begin{tabular}{|l|c|}
\hline
Recall & 53.23\% \\
Precision & 59.37\% \\
Hmean & 56.13\% \\
\hline\hline
1-N.E.D & 63.16\% \\
\hline
\end{tabular}
\end{center}
\caption{STR Results on ICDAR2019-LSVT.}
\label{table:result_lsvt}
\end{table}

\begin{table}
\begin{center}
\begin{tabular}{|l|c|}
\hline
Recall & 93.62\% \\
Precision & 87.22\% \\
Hmean & 90.30\% \\
\hline\hline
1-N.E.D & 76.60\% \\
\hline
\end{tabular}
\end{center}
\caption{STR Results on ICDAR2019-ReCTS.}
\label{table:result_rects}
\end{table}

Fig. \ref{fig:viz} shows the text detection and recognition results. As reported in ~\cite{chng2019icdar2019}, our algorithm achieves the championship in the task of scene text spotting with the accuracy H-mean score of 52.45\% in Latin and 50.17\% in Latin and Chinese. Normalized Edit Distance metric (1-N.E.D specifically) is treated as the official ranking metric is formulated as below:
\begin{eqnarray}
Norm = 1 - \frac{1}{N}\sum_{1}^{N}D(s_{i}, \hat{s}_{i})/max(s_{i}, \hat{s}_{i})
\end{eqnarray}
where $D(:)$ stands for the Levenshtein Distance, and $s_{i}$ and $\hat{s}_{i}$ denote the predicted text line in string and the corresponding ground truths in the regions. Note that the results are tested in an unique model. Numeric results on Latin Only, Latin and Chinese are listed in Table \ref{table:result_latin} and Table \ref{table:result_latin_and_chinese}, respectively. Moreover, we also achieve competitive results on ICDAR2019-LSVT in Table \ref{table:result_lsvt} and ICDAR2019-ReCTS in Table \ref{table:result_rects}, which demonstrate the effectiveness and robustness of our STR algorithm. Note that the detection model is trained on different datasets seperately with pretrained detection model from ICDAR2017-MLT and ICDAR2017RCTW, while the recognition model is the same one as mentioned in Section 4.2.

\section{Conclusion and Future Work}

\begin{figure}[t]
\begin{center}
\includegraphics[width=0.9\linewidth]{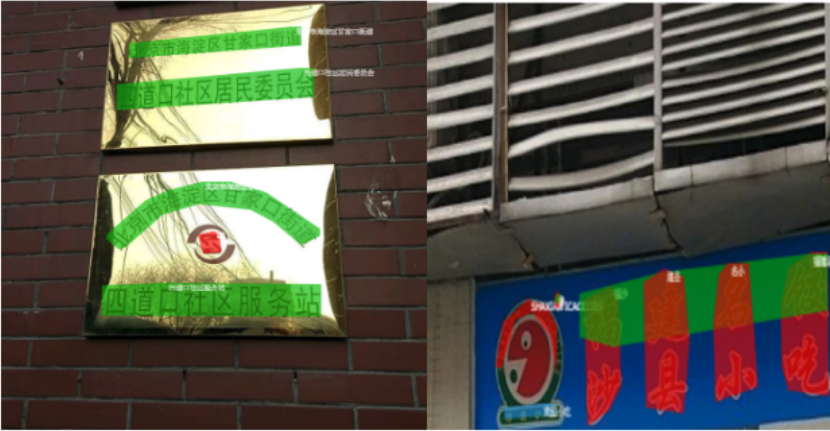}
\end{center}
   \caption{Failure case. The failed case indicates the necessity of end-to-end STR that recognition might contribute to detection.}
\label{fig:failure}
\end{figure}

This paper aims to propose a universal framework for STR with the combination of object detection and attention based language model. The competition results in ICDAR2019 including LSVT, ArT, ReCTS demonstrate the effectiveness and robustness of our algorithm. In addition, it is convenient to extend the current algorithm to state-of-the-art method such as replacing detection or attention mechanism with better architectures, such as fast and accurate detector EfficientDet~\cite{tan2019efficientdet} or Transformer~\cite{vaswani2017attention} based self-attention mechanism. Moreover, it is feasible to synthesize text images from natural language process corpus for data augmentation and helpful for attention-based language model.

In future, it is imperative to build an end-to-end differentiable STR algorithm with both detection and recognition which requires GPUs with large memory like V100. It is essential to eliminate the detection failure case as illustrated in Fig. \ref{fig:failure} with semantic information. Language model based methods like BERT~\cite{devlin2018bert} should be beneficial in our framework which takes the context of whole sentences into consideration instead of the previous word merely. Furthermore, visual based question answering or semantic analysis modules can be integrated with the framework for text based high-level semantic applications.


{\small
\bibliographystyle{ieee_fullname}
\bibliography{egbib}

\begin{thebibliography}{10}\itemsep=-1pt

\bibitem{abadi2016tensorflow}
Mart{\'\i}n Abadi, Paul Barham, Jianmin Chen, Zhifeng Chen, Andy Davis, Jeffrey
  Dean, Matthieu Devin, Sanjay Ghemawat, Geoffrey Irving, Michael Isard, et~al.
\newblock Tensorflow: A system for large-scale machine learning.
\newblock In {\em 12th $\{$USENIX$\}$ Symposium on Operating Systems Design and
  Implementation ($\{$OSDI$\}$ 16)}, pages 265--283, 2016.

\bibitem{bahdanau2014neural}
Dzmitry Bahdanau, Kyunghyun Cho, and Yoshua Bengio.
\newblock Neural machine translation by jointly learning to align and
  translate.
\newblock {\em arXiv preprint arXiv:1409.0473}, 2014.

\bibitem{cai2018cascade}
Zhaowei Cai and Nuno Vasconcelos.
\newblock Cascade r-cnn: Delving into high quality object detection.
\newblock In {\em Proceedings of the IEEE conference on computer vision and
  pattern recognition}, pages 6154--6162, 2018.

\bibitem{cheng2016long}
Jianpeng Cheng, Li Dong, and Mirella Lapata.
\newblock Long short-term memory-networks for machine reading.
\newblock {\em arXiv preprint arXiv:1601.06733}, 2016.

\bibitem{chng2019icdar2019}
Chee-Kheng Chng, Yuliang Liu, Yipeng Sun, Chun~Chet Ng, Canjie Luo, Zihan Ni,
  ChuanMing Fang, Shuaitao Zhang, Junyu Han, Errui Ding, et~al.
\newblock Icdar2019 robust reading challenge on arbitrary-shaped text
  (rrc-art).
\newblock {\em arXiv preprint arXiv:1909.07145}, 2019.

\bibitem{dai2017deformable}
Jifeng Dai, Haozhi Qi, Yuwen Xiong, Yi Li, Guodong Zhang, Han Hu, and Yichen
  Wei.
\newblock Deformable convolutional networks.
\newblock In {\em Proceedings of the IEEE international conference on computer
  vision}, pages 764--773, 2017.

\bibitem{deng2009imagenet}
Jia Deng, Wei Dong, Richard Socher, Li-Jia Li, Kai Li, and Li Fei-Fei.
\newblock Imagenet: A large-scale hierarchical image database.
\newblock In {\em 2009 IEEE conference on computer vision and pattern
  recognition}, pages 248--255. Ieee, 2009.

\bibitem{devlin2018bert}
Jacob Devlin, Ming-Wei Chang, Kenton Lee, and Kristina Toutanova.
\newblock Bert: Pre-training of deep bidirectional transformers for language
  understanding.
\newblock {\em arXiv preprint arXiv:1810.04805}, 2018.

\bibitem{duan2019centernet}
Kaiwen Duan, Song Bai, Lingxi Xie, Honggang Qi, Qingming Huang, and Qi Tian.
\newblock Centernet: Keypoint triplets for object detection.
\newblock In {\em Proceedings of the IEEE International Conference on Computer
  Vision}, pages 6569--6578, 2019.

\bibitem{gehring2017convolutional}
Jonas Gehring, Michael Auli, David Grangier, Denis Yarats, and Yann~N Dauphin.
\newblock Convolutional sequence to sequence learning.
\newblock In {\em Proceedings of the 34th International Conference on Machine
  Learning-Volume 70}, pages 1243--1252. JMLR. org, 2017.

\bibitem{ghiasi2019fpn}
Golnaz Ghiasi, Tsung-Yi Lin, and Quoc~V Le.
\newblock Nas-fpn: Learning scalable feature pyramid architecture for object
  detection.
\newblock In {\em Proceedings of the IEEE Conference on Computer Vision and
  Pattern Recognition}, pages 7036--7045, 2019.

\bibitem{girshick2015fast}
Ross Girshick.
\newblock Fast r-cnn.
\newblock In {\em Proceedings of the IEEE international conference on computer
  vision}, pages 1440--1448, 2015.

\bibitem{girshick2014rich}
Ross Girshick, Jeff Donahue, Trevor Darrell, and Jitendra Malik.
\newblock Rich feature hierarchies for accurate object detection and semantic
  segmentation.
\newblock In {\em Proceedings of the IEEE conference on computer vision and
  pattern recognition}, pages 580--587, 2014.

\bibitem{graves2006connectionist}
Alex Graves, Santiago Fern{\'a}ndez, Faustino Gomez, and J{\"u}rgen
  Schmidhuber.
\newblock Connectionist temporal classification: labelling unsegmented sequence
  data with recurrent neural networks.
\newblock In {\em Proceedings of the 23rd international conference on Machine
  learning}, pages 369--376. ACM, 2006.

\bibitem{he2017mask}
Kaiming He, Georgia Gkioxari, Piotr Doll{\'a}r, and Ross Girshick.
\newblock Mask r-cnn.
\newblock In {\em Proceedings of the IEEE international conference on computer
  vision}, pages 2961--2969, 2017.

\bibitem{he2016deep}
Kaiming He, Xiangyu Zhang, Shaoqing Ren, and Jian Sun.
\newblock Deep residual learning for image recognition.
\newblock In {\em Proceedings of the IEEE conference on computer vision and
  pattern recognition}, pages 770--778, 2016.

\bibitem{hochreiter1997long}
Sepp Hochreiter and J{\"u}rgen Schmidhuber.
\newblock Long short-term memory.
\newblock {\em Neural computation}, 9(8):1735--1780, 1997.

\bibitem{hu2018squeeze}
Jie Hu, Li Shen, and Gang Sun.
\newblock Squeeze-and-excitation networks.
\newblock In {\em Proceedings of the IEEE conference on computer vision and
  pattern recognition}, pages 7132--7141, 2018.

\bibitem{ioffe2015batch}
Sergey Ioffe and Christian Szegedy.
\newblock Batch normalization: Accelerating deep network training by reducing
  internal covariate shift.
\newblock {\em arXiv preprint arXiv:1502.03167}, 2015.

\bibitem{kalchbrenner2016neural}
Nal Kalchbrenner, Lasse Espeholt, Karen Simonyan, Aaron van~den Oord, Alex
  Graves, and Koray Kavukcuoglu.
\newblock Neural machine translation in linear time.
\newblock {\em arXiv preprint arXiv:1610.10099}, 2016.

\bibitem{kingma2014adam}
Diederik~P Kingma and Jimmy Ba.
\newblock Adam: A method for stochastic optimization.
\newblock {\em arXiv preprint arXiv:1412.6980}, 2014.

\bibitem{krizhevsky2012imagenet}
Alex Krizhevsky, Ilya Sutskever, and Geoffrey~E Hinton.
\newblock Imagenet classification with deep convolutional neural networks.
\newblock In {\em Advances in neural information processing systems}, pages
  1097--1105, 2012.

\bibitem{law2018cornernet}
Hei Law and Jia Deng.
\newblock Cornernet: Detecting objects as paired keypoints.
\newblock In {\em Proceedings of the European Conference on Computer Vision
  (ECCV)}, pages 734--750, 2018.

\bibitem{LiaoSBWL17}
Minghui Liao, Baoguang Shi, Xiang Bai, Xinggang Wang, and Wenyu Liu.
\newblock Textboxes: {A} fast text detector with a single deep neural network.
\newblock In {\em AAAI}, 2017.

\bibitem{lin2017feature}
Tsung-Yi Lin, Piotr Doll{\'a}r, Ross Girshick, Kaiming He, Bharath Hariharan,
  and Serge Belongie.
\newblock Feature pyramid networks for object detection.
\newblock In {\em Proceedings of the IEEE conference on computer vision and
  pattern recognition}, pages 2117--2125, 2017.

\bibitem{lin2017focal}
Tsung-Yi Lin, Priya Goyal, Ross Girshick, Kaiming He, and Piotr Doll{\'a}r.
\newblock Focal loss for dense object detection.
\newblock In {\em Proceedings of the IEEE international conference on computer
  vision}, pages 2980--2988, 2017.

\bibitem{liu2018progressive}
Chenxi Liu, Barret Zoph, Maxim Neumann, Jonathon Shlens, Wei Hua, Li-Jia Li, Li
  Fei-Fei, Alan Yuille, Jonathan Huang, and Kevin Murphy.
\newblock Progressive neural architecture search.
\newblock In {\em Proceedings of the European Conference on Computer Vision
  (ECCV)}, pages 19--34, 2018.

\bibitem{liu2016ssd}
Wei Liu, Dragomir Anguelov, Dumitru Erhan, Christian Szegedy, Scott Reed,
  Cheng-Yang Fu, and Alexander~C Berg.
\newblock Ssd: Single shot multibox detector.
\newblock In {\em European conference on computer vision}, pages 21--37.
  Springer, 2016.

\bibitem{liu2018fots}
Xuebo Liu, Ding Liang, Shi Yan, Dagui Chen, Yu Qiao, and Junjie Yan.
\newblock Fots: Fast oriented text spotting with a unified network.
\newblock In {\em Proceedings of the IEEE conference on computer vision and
  pattern recognition}, pages 5676--5685, 2018.

\bibitem{luong17}
Minh{-}Thang Luong, Eugene Brevdo, and Rui Zhao.
\newblock Neural machine translation (seq2seq) tutorial.
\newblock {\em https://github.com/tensorflow/nmt}, 2017.

\bibitem{luong2015effective}
Minh-Thang Luong, Hieu Pham, and Christopher~D Manning.
\newblock Effective approaches to attention-based neural machine translation.
\newblock {\em arXiv preprint arXiv:1508.04025}, 2015.

\bibitem{lyu2018mask}
Pengyuan Lyu, Minghui Liao, Cong Yao, Wenhao Wu, and Xiang Bai.
\newblock Mask textspotter: An end-to-end trainable neural network for spotting
  text with arbitrary shapes.
\newblock In {\em Proceedings of the European Conference on Computer Vision
  (ECCV)}, pages 67--83, 2018.

\bibitem{mikolov2013efficient}
Tomas Mikolov, Kai Chen, Greg Corrado, and Jeffrey Dean.
\newblock Efficient estimation of word representations in vector space.
\newblock {\em arXiv preprint arXiv:1301.3781}, 2013.

\bibitem{Liao2018Text}
Baoguang~Shi Minghui~Liao and Xiang Bai.
\newblock {TextBoxes++}: A single-shot oriented scene text detector.
\newblock {\em {IEEE} Transactions on Image Processing}, 27(8):3676--3690,
  2018.

\bibitem{mnih2014recurrent}
Volodymyr Mnih, Nicolas Heess, Alex Graves, et~al.
\newblock Recurrent models of visual attention.
\newblock In {\em Advances in neural information processing systems}, pages
  2204--2212, 2014.

\bibitem{nayef2017icdar2017}
Nibal Nayef, Fei Yin, Imen Bizid, Hyunsoo Choi, Yuan Feng, Dimosthenis
  Karatzas, Zhenbo Luo, Umapada Pal, Christophe Rigaud, Joseph Chazalon, et~al.
\newblock Icdar2017 robust reading challenge on multi-lingual scene text
  detection and script identification-rrc-mlt.
\newblock In {\em 2017 14th IAPR International Conference on Document Analysis
  and Recognition (ICDAR)}, volume~1, pages 1454--1459. IEEE, 2017.

\bibitem{redmon2016you}
Joseph Redmon, Santosh Divvala, Ross Girshick, and Ali Farhadi.
\newblock You only look once: Unified, real-time object detection.
\newblock In {\em Proceedings of the IEEE conference on computer vision and
  pattern recognition}, pages 779--788, 2016.

\bibitem{redmon2017yolo9000}
Joseph Redmon and Ali Farhadi.
\newblock Yolo9000: better, faster, stronger.
\newblock In {\em Proceedings of the IEEE conference on computer vision and
  pattern recognition}, pages 7263--7271, 2017.

\bibitem{redmon2018yolov3}
Joseph Redmon and Ali Farhadi.
\newblock Yolov3: An incremental improvement.
\newblock {\em arXiv preprint arXiv:1804.02767}, 2018.

\bibitem{ren2015faster}
Shaoqing Ren, Kaiming He, Ross Girshick, and Jian Sun.
\newblock Faster r-cnn: Towards real-time object detection with region proposal
  networks.
\newblock In {\em Advances in neural information processing systems}, pages
  91--99, 2015.

\bibitem{shi2016end}
Baoguang Shi, Xiang Bai, and Cong Yao.
\newblock An end-to-end trainable neural network for image-based sequence
  recognition and its application to scene text recognition.
\newblock {\em IEEE transactions on pattern analysis and machine intelligence},
  39(11):2298--2304, 2016.

\bibitem{shi2017icdar2017}
Baoguang Shi, Cong Yao, Minghui Liao, Mingkun Yang, Pei Xu, Linyan Cui, Serge
  Belongie, Shijian Lu, and Xiang Bai.
\newblock Icdar2017 competition on reading chinese text in the wild (rctw-17).
\newblock In {\em 2017 14th IAPR International Conference on Document Analysis
  and Recognition (ICDAR)}, volume~1, pages 1429--1434. IEEE, 2017.

\bibitem{singh2018analysis}
Bharat Singh and Larry~S Davis.
\newblock An analysis of scale invariance in object detection snip.
\newblock In {\em Proceedings of the IEEE conference on computer vision and
  pattern recognition}, pages 3578--3587, 2018.

\bibitem{singh2018sniper}
Bharat Singh, Mahyar Najibi, and Larry~S Davis.
\newblock Sniper: Efficient multi-scale training.
\newblock In {\em Advances in Neural Information Processing Systems}, pages
  9310--9320, 2018.

\bibitem{sukhbaatar2015end}
Sainbayar Sukhbaatar, Jason Weston, Rob Fergus, et~al.
\newblock End-to-end memory networks.
\newblock In {\em Advances in neural information processing systems}, pages
  2440--2448, 2015.

\bibitem{sutskever2014sequence}
I Sutskever, O Vinyals, and QV Le.
\newblock Sequence to sequence learning with neural networks.
\newblock {\em Advances in NIPS}, 2014.

\bibitem{szegedy2017inception}
Christian Szegedy, Sergey Ioffe, Vincent Vanhoucke, and Alexander~A Alemi.
\newblock Inception-v4, inception-resnet and the impact of residual connections
  on learning.
\newblock In {\em Thirty-First AAAI Conference on Artificial Intelligence},
  2017.

\bibitem{szegedy2016rethinking}
Christian Szegedy, Vincent Vanhoucke, Sergey Ioffe, Jon Shlens, and Zbigniew
  Wojna.
\newblock Rethinking the inception architecture for computer vision.
\newblock In {\em Proceedings of the IEEE conference on computer vision and
  pattern recognition}, pages 2818--2826, 2016.

\bibitem{tan2019efficientnet}
Mingxing Tan and Quoc~V Le.
\newblock Efficientnet: Rethinking model scaling for convolutional neural
  networks.
\newblock {\em arXiv preprint arXiv:1905.11946}, 2019.

\bibitem{tan2019efficientdet}
Mingxing Tan, Ruoming Pang, and Quoc~V Le.
\newblock Efficientdet: Scalable and efficient object detection.
\newblock {\em arXiv preprint arXiv:1911.09070}, 2019.

\bibitem{tian2019fcos}
Zhi Tian, Chunhua Shen, Hao Chen, and Tong He.
\newblock Fcos: Fully convolutional one-stage object detection.
\newblock {\em arXiv preprint arXiv:1904.01355}, 2019.

\bibitem{vaswani2017attention}
Ashish Vaswani, Noam Shazeer, Niki Parmar, Jakob Uszkoreit, Llion Jones,
  Aidan~N Gomez, {\L}ukasz Kaiser, and Illia Polosukhin.
\newblock Attention is all you need.
\newblock In {\em Advances in neural information processing systems}, pages
  5998--6008, 2017.

\bibitem{wu2016tensorpack}
Yuxin Wu et~al.
\newblock Tensorpack.
\newblock \url{https://github.com/tensorpack/}, 2016.

\bibitem{wu2016google}
Yonghui Wu, Mike Schuster, Zhifeng Chen, Quoc~V Le, Mohammad Norouzi, Wolfgang
  Macherey, Maxim Krikun, Yuan Cao, Qin Gao, Klaus Macherey, et~al.
\newblock Google's neural machine translation system: Bridging the gap between
  human and machine translation.
\newblock {\em arXiv preprint arXiv:1609.08144}, 2016.

\bibitem{xu2015show}
Kelvin Xu, Jimmy Ba, Ryan Kiros, Kyunghyun Cho, Aaron Courville, Ruslan
  Salakhudinov, Rich Zemel, and Yoshua Bengio.
\newblock Show, attend and tell: Neural image caption generation with visual
  attention.
\newblock In {\em International conference on machine learning}, pages
  2048--2057, 2015.

\bibitem{zhou2017east}
Xinyu Zhou, Cong Yao, He Wen, Yuzhi Wang, Shuchang Zhou, Weiran He, and Jiajun
  Liang.
\newblock East: an efficient and accurate scene text detector.
\newblock In {\em Proceedings of the IEEE conference on Computer Vision and
  Pattern Recognition}, pages 5551--5560, 2017.

\bibitem{zhu2019deformable}
Xizhou Zhu, Han Hu, Stephen Lin, and Jifeng Dai.
\newblock Deformable convnets v2: More deformable, better results.
\newblock In {\em Proceedings of the IEEE Conference on Computer Vision and
  Pattern Recognition}, pages 9308--9316, 2019.

\bibitem{zoph2016neural}
Barret Zoph and Quoc~V Le.
\newblock Neural architecture search with reinforcement learning.
\newblock {\em arXiv preprint arXiv:1611.01578}, 2016.

\end{thebibliography}
}

\end{document}